\pdfoutput=1

\documentclass[11pt]{article}

\usepackage{ACL2023}
\usepackage{todonotes}
\usepackage{times}
\usepackage{latexsym}
\usepackage{xcolor}

\usepackage{booktabs}
\usepackage{multirow}
\usepackage{graphicx}
\usepackage{adjustbox}
\usepackage{xcolor}

\usepackage[T1]{fontenc}

\usepackage[utf8]{inputenc}

\usepackage{microtype}

\usepackage{inconsolata}

\newcommand{\ignore}[1]{}

\title{When Semantic Overlap Is Not Enough: Cross-Lingual Euphemism Transfer Between Turkish and English}

\author{
Hasan Can Biyik\textsuperscript{*} \quad
Libby Barak \quad
Jing Peng \quad
Anna Feldman \\[0.5em]
Montclair State University \\
\texttt{\{biyikh1, barakl, pengj, feldmana\}@montclair.edu} \\[0.3em]
}

\begin{document}
\maketitle
\begin{abstract}
Euphemisms substitute socially sensitive expressions, often softening or reframing meaning, and their reliance on cultural and pragmatic context complicates modeling across languages. In this study, we investigate how cross-lingual equivalence influences transfer in multilingual euphemism detection. We categorize Potentially Euphemistic Terms (PETs) in Turkish and English into Overlapping (OPETs) and Non-Overlapping (NOPETs) subsets based on their functional, pragmatic, and semantic alignment.  Our findings reveal a transfer asymmetry: semantic overlap is insufficient to guarantee positive transfer, particularly in low-resource Turkish-to-English direction, where performance can degrade even for overlapping euphemisms, and in some cases, improve under NOPET-based training. Differences in label distribution help explain these counterintuitive results. Category-level analysis suggests that transfer may be influenced by domain-specific alignment, though evidence is limited by sparsity.
\end{abstract}


\section{Introduction}
Euphemisms serve a critical pragmatic function: they soften harsh, impolite, or taboo expressions while preserving their intended meaning. For instance, instead of stating that someone was \emph{fired}, one might describe them as \emph{let go}. While the intent remains constant, the realization of euphemisms is deeply figurative and culture-dependent, making computational detection challenging \cite{gavidia-2022-cats}.

\citet{lee-etal-2022-searching} formalized this task by introducing Potentially Euphemistic Terms (PETs). While euphemisms exist universally, their specific forms vary significantly across cultures \cite{lee-etal-2024-meds}. This creates a unique challenge for multilingual language models: can a model learn the \textit{concept} of a euphemism in one language and apply it to another, or is detection strictly bound by cultural familiarity?

Beyond its theoretical relevance, cross-lingual euphemism detection has practical implications for several NLP applications. Euphemisms are frequently used to obscure sensitive or policy-violating content in online platforms for content moderation and hate speech monitoring.

In this study, we investigate the limits of cross-lingual transfer in  XLM-RoBERTa (XLM-R) \cite{conneau-etal-2020-unsupervised}. To examine  how cross-lingual semantic equivalence influences transfer, we categorize PETs into two distinct subsets based on their functional, pragmatic, and  semantic alignment across languages: \textbf{Overlapping PETs (OPETs)}, which share functional and semantic equivalents across languages  (e.g., \textit{passed away} $\approx$ \textit{vefat etmek}), and  \textbf{Non-Overlapping PETs (NOPETs)}, which lack functional or  semantic equivalents in the target language.

Previous work has explored PET disambiguation in multilingual contexts \cite{lee-etal-2024-meds}, yet the specific impact of semantic overlap on zero-shot transfer remains underexplored. We bridge this gap by expanding the existing Turkish PETs dataset to create aligned OPET/NOPET subsets, allowing us to examine how semantic overlap interacts with zero-shot transfer in euphemism detection, and to diagnose when cross-lingual generalization fails despite shared multilingual representations.

Our contributions are as follows:
\begin{itemize}
    \setlength{\itemsep}{0pt}
    \setlength{\parskip}{0pt}
    \item We introduce the first cross-lingual euphemism dataset explicitly categorized into overlapping (OPET) and non-overlapping (NOPET) subsets for English and Turkish.
    \item We reveal a transfer asymmetry most pronounced at the category level: while English-to-Turkish transfer remains robust across domains (e.g., Employment F1=0.90, Death F1=0.86), Turkish-to-English transfer degrades substantially in certain categories (Employment F1=0.36, Politics F1=0.18).
    \item We provide detailed OPET/NOPET annotation guidelines and release the code and data splits for all experiments.\footnote{Code and data are publicly available at \url{https://github.com/hasancanbiyik/PETs-investigating-LLMs-generalization-capabilities}.}
\end{itemize}


\section{Related Work}
\paragraph{Euphemism Detection.}
Early computational approaches to euphemism detection were predominantly English-centric and lexicon-based, relying on sentiment analysis and  handcrafted semantic cues \cite{felt-riloff-2020-recognizing,  magu-luo-2018-determining}. With the advent of deep learning, the field shifted toward transformer-based models that leverage masked language modeling for superior contextual understanding \cite{zhu-2021-self-supervised, kapron-king-xu-2021-diachronic}. To systematically address the variability of euphemistic usage, \citet{gavidia-2022-cats} introduced the concept of Potentially Euphemistic Terms (PETs), a framework later expanded by \citet{lee-etal-2022-searching} and enhanced by linguistic feature  integration \cite{lee-etal-2024-meds}. Recent work has investigated cross-lingual transfer for euphemism detection, with \citet{sammartino2025ranlp} showing that sequential fine-tuning (training on L1 before L2) outperforms simultaneous multilingual training for low-resource languages, though it suffers from catastrophic forgetting in models with uneven pretraining coverage like XLM-R.

\paragraph{Multilingual Resources.}
Building on these detection frameworks, recent shared tasks have introduced euphemism detection in English \cite{lee-etal-2022-report} and subsequently expanded to multilingual settings \cite{lee-feldman-2024-report}. New PET datasets have emerged for Mandarin Chinese, Spanish, and Yorùbá \cite{lee-etal-2024-meds}, as well as Danish \cite{al-laith-etal-2025-dying} and Turkish \cite{biyik-etal-2024-turkish}. However, they do not explicitly explore the cross-lingual transfer, meaning how \textit{semantic and functional overlap} (or the lack thereof) facilitates or inhibits euphemism detection across languages.

\paragraph{Cross-Lingual Transfer in NLP.}
Multilingual masked language models like XLM-R have become the standard for zero-shot cross-lingual transfer, consistently outperforming previous architectures like mBERT due to larger vocabularies and shared embedding spaces \cite{wu-dredze-2019-beto}. While XLM-R has shown strong performance on syntactic and literal semantic tasks, transferring figurative knowledge remains a significant challenge. Our work addresses this gap by investigating XLM-R's ability to transfer euphemism knowledge between Turkish and English, languages with distinct typological and cultural characteristics.

\paragraph{Linguistic and Sociolinguistic Perspectives.}
Outside of NLP, euphemisms have been primarily studied in linguistics and sociolinguistics as pragmatic and culturally situated phenomena, with foundational work focusing on taboo, politeness, and social meaning rather than contextual disambiguation or prediction \cite{allan2006forbidden,burridge}. This literature is relatively limited and largely descriptive, offering taxonomies and sociocultural analyses but little guidance for operationalizing euphemism detection, particularly in multilingual or cross-lingual settings. The scarcity of formal linguistic models for euphemism interpretation helps explain why euphemism detection remains challenging for computational systems, especially when meanings rely on shared cultural knowledge rather than lexical cues alone.


\section{Datasets}
This study uses PETs datasets in English and Turkish, categorized into domains such as death, sexual activity, physical/mental attributes, politics, body functions, employment, and illness. For example, \emph{go to heaven} euphemistically means \emph{to die}, belonging to the death category. Due to the original dataset being too small, we expand the Turkish PETs dataset to have similar numbers of euphemistic terms and examples. We then identify the OPETs and NOPETs between the languages.

\subsection{Existing Datasets}
The English PETs dataset was introduced by \citet{gavidia-2022-cats}; Mandarin Chinese, Spanish, and Yorùbá PETs datasets were later presented by \citet{lee-etal-2024-meds}. The initial Turkish dataset was presented by \citet{biyik-etal-2024-turkish}. Dataset statistics are summarized in Table~\ref{tbl:all_pets_all_languages}.

\begin{table}[h]
\centering
\small
\setlength{\tabcolsep}{3pt}
\renewcommand{\arraystretch}{0.85}
\begin{tabular}{lccccc}
    \toprule
    & \textbf{EN} & \textbf{TR} & \textbf{ZH} & \textbf{ES} & \textbf{YO} \\
    \midrule
    PETs & 144 & 70 & 149 & 233 & 157 \\
    Total & 3098 & 2436 & 3211 & 2952 & 2598 \\
    Euphemistic & 1841 & 1457 & 2213 & 1955 & 1689 \\
    Non-Euph. & 1257 & 979 & 998 & 997 & 909 \\
    \bottomrule
\end{tabular}
\caption{Statistics of PETs datasets across languages. The Turkish (TR) column reflects our newly expanded dataset (70 PETs). For English, we use a subset of 71 PETs from the original 144 PETs to create a balanced sample for both languages.}
\label{tbl:all_pets_all_languages}
\end{table}

To illustrate the distinction between euphemistic and non-euphemistic uses, consider the PET \emph{disabled}, an example taken from the English PETs dataset \cite{gavidia-2022-cats}:

\noindent
\textbf{Euphemistic Use:} \emph{I am a 40 year old <disabled> vet who has returned to school for my political science degree.}

\noindent
\textbf{Non-euphemistic Use:} \emph{The firewall and av are <disabled>.}

\subsection{Expansion of the Turkish PETs Dataset}\label{sec:turkishdataset}
The original resource created by \citeauthor{biyik-etal-2024-turkish} contains over 120 PETs with associated metadata (e.g., substituted term, semantic category, non-euphemistic meaning, euphemistic meaning, and source). We were able to reliably collect sufficient contextual examples for approximately 30 PETs out of this list bringing the total number of PETs in our data to 70 PETs.

The original 40 PETs already present in the dataset were retained without modification; for these PETs, we only adjusted the number of instances to achieve a more balanced distribution between euphemistic and non-euphemistic usages.

We expanded the dataset to 70 PETs by adding potentially euphemistic terms from various semantic categories (e.g., death, employment, sexual activity) and collecting examples from SketchEngine's Turkish Web 2020 Corpus (trTenTen20) \citep{kilgarriff2014sketch}. We evaluated the impact of the proposed expansion.

Table~\ref{tbl:old_vs_new_f1} compares the overall performance before and after. The expanded dataset introduced lower-frequency and more diverse euphemisms, creating a more challenging and realistic benchmark. Model performance remained largely stable, with F1 scores changing by at most 0.04 points, validating the quality of the expansion.

\begin{table}[h]
\centering
\small
\setlength{\tabcolsep}{8pt}
\renewcommand{\arraystretch}{1.0}
\begin{tabular}{lcc}
    \toprule
    \textbf{Model} & \textbf{Previous F1} & \textbf{New F1} \\
    \midrule
    BERTurk & 0.84 & 0.84 \\
    Electra & 0.86 & 0.83 \\
    mBERT & 0.80 & 0.76 \\
    XLM-R & 0.82 & 0.81 \\
    \bottomrule
\end{tabular}
\caption{Comparison of old vs. new F1 scores after expanding the Turkish PETs dataset.}
\label{tbl:old_vs_new_f1}
\end{table}

\subsection{Overlap Annotation and Criteria}
\label{sec:overlap_criteria}
A core contribution of this work is the rigorous categorization of PETs into Overlapping (OPETs) and Non-Overlapping (NOPETs) subsets. While \citet{biyik-etal-2024-turkish} briefly noted the existence of cross-lingual overlap, we formalize this distinction through a systematic annotation process that defines cross-lingual overlap based on functional and semantic equivalence rather than literal translation.

\paragraph{Annotation Framework}
We define cross-lingual overlap based on \textbf{functional equivalence}: a PET is classified as an OPET if both languages possess expressions that (1) address the same underlying taboo concept and (2) employ a similar euphemistic strategy (e.g., both frame death as departure). This definition prioritizes pragmatic function over lexical similarity. For a comprehensive overview of the decision process and additional examples, we refer the reader to the full annotation guidelines in Appendix~\ref{sec:appendix_guidelines}.

\begin{itemize}
    \item \textbf{Overlapping PETs (OPETs):} Terms with a functional euphemistic equivalent in the target language. For example, \textit{comfort woman} and \textit{hayat kadını} are OPETs because they both function to soften the concept of ``prostitute'', despite lacking lexical overlap. Similarly, \textit{pass away} and \textit{vefat etmek} both euphemize death through departure framing.
    \item \textbf{Non-Overlapping PETs (NOPETs):} Terms where the euphemistic mapping may exist in the source language but lacks a functional equivalent in the target language. For instance, the English PET \textit{the birds and the bees} (sex education) has no functional equivalent in Turkish and is therefore labeled as a NOPET.
\end{itemize}

\paragraph{Reliability}
To ensure the quality of these subsets, we measured Inter-Annotator Agreement (IAA) using Cohen's Kappa ($\kappa$). The annotators achieved a score of $\kappa = 0.96$ (Raw Agreement: 96\%), indicating near-perfect reliability. This high agreement is partly attributable to the binary nature of the task and the use of detailed annotation guidelines. Disagreements were resolved through adjudication. Table~\ref{tab:opet_nopet_and_categories} details the final distribution of these subsets.

\begin{table*}[t]
\centering
\small
\begin{minipage}[t]{0.52\textwidth}
\centering
\setlength{\tabcolsep}{5pt}
\renewcommand{\arraystretch}{1.1}
\begin{tabular}{lcccc}
\toprule
 & \multicolumn{2}{c}{\textbf{Turkish}} & \multicolumn{2}{c}{\textbf{English}} \\
\cmidrule(lr){2-3} \cmidrule(lr){4-5}
 & OPET & NOPET & OPET & NOPET \\
\midrule
PETs & 36 & 34 & 36 & 35 \\
Instances & 1130 & 1002 & 1113 & 973 \\
Euphemistic & 578 & 591 & 579 & 481 \\
Non-Euphemistic & 552 & 411 & 534 & 492 \\
Always Euph. & 7 & 4 & 13 & 0 \\
Never Euph. & 1 & 2 & 0 & 0 \\
\bottomrule
\end{tabular}
\caption*{(a) OPET/NOPET Distribution}
\end{minipage}%
\hspace{0.5cm}
\begin{minipage}[t]{0.44\textwidth}
\centering
\setlength{\tabcolsep}{6pt}
\renewcommand{\arraystretch}{1.1}
\begin{tabular}{lcc}
\toprule
\textbf{Category} & \textbf{TR} & \textbf{EN} \\
\midrule
Death & 21 (661) & 15 (580) \\
Body Functions & 14 (469) & 4 (32) \\
Sexual Activity & 8 (226) & 10 (265) \\
Employment & 7 (167) & 14 (430) \\
Phys./Ment. Attr. & 11 (257) & 19 (563) \\
Politics & 1 (11) & 3 (83) \\
Illness & 2 (65) & 1 (20) \\
Substances & 1 (80) & 4 (120) \\
Miscellaneous & 5 (198) & 1 (32) \\
\bottomrule
\end{tabular}
\caption*{(b) Category Distribution}
\end{minipage}
\caption{Distribution of PETs across OPET/NOPET subsets (a) and semantic categories (b). Category values show number of PETs (number of instances). Terms marked "Always Euph." are used exclusively in euphemistic contexts; "Never Euph." appear only in literal uses.}
\label{tab:opet_nopet_and_categories}
\end{table*}


\section{Methodology}
We conduct a series of experiments to evaluate XLM-R’s euphemism detection across OPET and NOPET subsets in Turkish and English, assessing both within-language generalization and cross-lingual transfer. We use XLM-RoBERTa (XLM-R), a multilingual masked language model pretrained on 100 languages including both Turkish and English. XLM-R has demonstrated strong zero-shot cross-lingual transfer capabilities on various language tasks, making it well-suited for evaluating euphemism detection across typologically distinct languages.

\subsection{Baseline Performance}
\label{ssec:baseline}
To assess whether pre-trained representations alone can distinguish euphemisms, we evaluate frozen XLM-R embeddings without task-specific fine-tuning. We extract contextualized representations of the target PET by mean-pooling the final-layer token embeddings corresponding to the PET span, and train a Logistic Regression (LR) classifier on these representations. As shown in Table~\ref{tab:baseline_performance}, OPETs consistently outperform NOPETs across both languages (e.g., 0.71 vs. 0.61 in English), suggesting that OPET classification is more tractable from frozen representations, possibly due to differences in how these terms are represented in the pretrained embedding space.

\begin{table}[ht]
    \centering
    \small
    \setlength{\tabcolsep}{6pt}
    \begin{tabular}{lcc|cc}
        \toprule
        & \multicolumn{2}{c|}{\textbf{English}} & \multicolumn{2}{c}{\textbf{Turkish}} \\
        \textbf{Metric} & \textbf{OPET} & \textbf{NOPET} & \textbf{OPET} & \textbf{NOPET} \\
        \midrule
        F1 Score & 0.71 & 0.61 & 0.72 & 0.68 \\
        \bottomrule
    \end{tabular}
    \caption{Baseline performance (F1) using frozen XLM-R embeddings.}
    \label{tab:baseline_performance}
\end{table}

\subsection{Experimental Setup}
We fine-tune XLM-R separately on each subset (English OPETs, English NOPETs, Turkish OPETs, Turkish NOPETs) using 10-fold cross-validation to ensure robust evaluation across diverse train-test-val data splits. Within each fold, we select the best checkpoint using a validation split drawn from the training portion of that fold.

\paragraph{Implementation Details.} 
We maintain consistent hyperparameters across all experiments: a learning rate of 1e-5, batch size of 4, and a maximum of 30 epochs with early stopping (patience=5). We do not use warmup steps or freeze any layers during fine-tuning.

\paragraph{Evaluation Protocol.} 
We evaluate cross-lingual transfer using a zero-shot protocol. A model fine-tuned on a source language subset (e.g., English OPETs) is evaluated directly on target language subsets (e.g., Turkish OPETs) without further training.
This assesses the robustness of the learned representations: high performance on the target subset indicates that the model successfully aligned euphemistic concepts across languages despite the language barrier.
Our cross-validation splits data at the instance level rather than grouping by PET, meaning the same PET may appear in both training and test sets across different instances. This could allow models to memorize specific terms rather than learn generalizable patterns. Future work should consider PET-grouped splits for stricter evaluation.


\section{Quantitative Results}
\label{sec:results}
We evaluate the fine-tuned XLM-R models using a direct zero-shot transfer protocol. Table~\ref{tab:fine_tuned_results} details the performance across all training and testing splits.

\subsection{Comparison with Baseline}
Fine-tuning yields substantial improvements over the frozen baseline when trained and tested on the same data. English OPET performance improves from 0.71 to 0.85 (+19.7\% relative gain), while Turkish NOPETs show substantial improvement, from 0.68 to 0.79 (+16.2\%). This indicates that fine-tuning is critical for euphemism detection regardless of overlap status.

\subsection{Cross-Lingual Transfer Analysis}
We observe two key trends supported by paired t-tests ($N=10$ folds).

\paragraph{Cross-Lingual Transfer vs Baseline.}
Comparing zero-shot transfer to the frozen baseline reveals a surprising pattern: while fine-tuning substantially improves in-domain performance, it can actually harm cross-lingual transfer. Training on English OPETs yields 0.68 F1 on Turkish OPETs, which is lower than the frozen baseline (0.72). This suggests that task-specific fine-tuning can reduce cross-lingual generalization, consistent with the hypothesis that fine-tuning emphasizes language-specific cues over language-universal patterns.

\paragraph{Does Semantic Overlap Constrain Transfer?}

A key research question is whether training on OPETs yields better cross-lingual transfer than training on NOPETs. We compare the training conditions for the same target data using the frozen baseline as a reference point.
When transferring to Turkish, training on English OPETs results in smaller degradation from baseline than training on English NOPETs. For the TR-OPET target, EN-OPET training yields 0.68 (a 0.04 drop from the 0.72 baseline), while EN-NOPET training yields 0.52 (a 0.20 drop). For the TR-NOPET target, EN-OPET training yields 0.70 (a 0.02 gain over the 0.68 baseline), while EN-NOPET training yields 0.56 (a 0.12 drop). This pattern suggests that OPET-trained models show smaller degradation from baseline in the EN→TR direction.

Critically, this pattern does not hold in the reverse direction. When transferring to English, both Turkish training conditions fall below the frozen baseline, with comparable drops: for the EN-OPET target, TR-OPET training yields 0.64 (a 0.07 drop from 0.71 baseline) and TR-NOPET training yields 0.66 (a 0.05 drop). For the EN-NOPET target, TR-OPET training yields 0.58 (a 0.03 drop from 0.61 baseline) and TR-NOPET training yields 0.60 (a 0.01 drop). This asymmetry suggests that OPET-trained models suffer smaller drops relative to baseline primarily when the source language is high-resource (English), but not when transferring from the low-resource language (Turkish).

The results present a nuanced picture. When training on English OPETs and testing on Turkish NOPETs, F1 is higher than the frozen baseline (0.70 compared with 0.68). However, performance on Turkish OPETs is actually slightly lower than its baseline when trained on English OPETs (0.68 vs 0.72). The importance of semantic overlap becomes evident when examining the reverse direction: training on Turkish OPETs leads to lower F1 for English OPETs and NOPETs compared with their frozen baselines. These performance gaps suggest that semantic overlap may matter more in the Turkish-to-English direction, possibly due to the familiarity of the model with language-specific cues prior to the fine-tuning stage. Previous work on cross-lingual transfer suggests that knowledge transfer might be sensitive for low-resource languages. Importantly, these results show that even semantic overlap does not reliably confer an advantage over non-overlapping training, which suggests that task-specific generalization may be bounded by baseline language proficiency.

Furthermore, comparing training conditions reveals that NOPET-trained models show larger drops from baseline than OPET-trained models, particularly in the EN→TR direction. Training on English NOPETs and testing on Turkish OPETs yields 0.52 compared with the 0.72 baseline (a 0.20 drop), while training on English OPETs yields 0.68 (only a 0.04 drop). This difference is consistent with our hypothesis that OPET-trained models degrade less from baseline than NOPET-trained models.

\paragraph{The Resource Asymmetry.}
Comparing transfer directions reveals a moderate asymmetry in NOPET performance. English-trained models transfer to Turkish NOPETs with F1 = 0.70, while Turkish-trained models transfer to English NOPETs with F1 = 0.58-0.60. While less dramatic than in-domain gaps, this difference is consistent with resource-imbalance effects in multilingual pretraining, though typological and morphological differences may also contribute. Overall transfer accuracy confirms this pattern: EN→TR transfers achieve 64--66\% accuracy, while TR→EN NOPET transfer drops to 56\%.

\begin{table*}[ht] 
    \centering
    \renewcommand{\arraystretch}{1.3} 
    \setlength{\tabcolsep}{0pt} 
    
\begin{tabular*}{\textwidth}{@{\extracolsep{\fill}} ll cc | cc }
    \toprule
    & & \multicolumn{2}{c|}{\textbf{Test on English}} & \multicolumn{2}{c}{\textbf{Test on Turkish}} \\
    \textbf{Train} & \textbf{Subset} & \textbf{OPET} & \textbf{NOPET} & \textbf{OPET} & \textbf{NOPET} \\
    \midrule 
    \textbf{Baseline}
    & (XLM-R)   & 0.71 & 0.61 & 0.72 & 0.68 \\ 
    \midrule
    \multirow{2}{*}{\textbf{EN}}  
    & OPET   & \textbf{0.85}* & 0.65 & 0.68 & 0.70 \\ 
    & NOPET  & 0.59 & \textbf{0.62}* & 0.52 & 0.56 \\ 
    \midrule
    \multirow{2}{*}{\textbf{TR}}  
    & OPET   & 0.64 & 0.58 & \textbf{0.80}* & 0.68 \\
    & NOPET  & 0.66 & 0.60 & 0.70 & \textbf{0.79}* \\
    \bottomrule
\end{tabular*}
    \caption{Cross-lingual transfer performance showing F1 scores and standard deviations across 10-fold cross-validation. Models are trained on source language subsets  (rows) and tested on target language subsets (columns). (*) indicates in-domain (same language, same subset) performance.}
    \label{tab:fine_tuned_results}
\end{table*}

\subsection{Category-Specific Performance}
To investigate whether certain semantic domains are more prone to transfer failure, we evaluated zero-shot performance across euphemism categories. Table~\ref{tab:category} presents F1 scores for key transfer conditions.
Because several semantic categories contain very few instances in specific OPET/NOPET subsets, the category-level results below should be interpreted as exploratory rather than definitive.

\begin{table}[t]
\centering
\small
\begin{tabular}{lcccc}
\toprule
& \multicolumn{2}{c}{\textbf{TR→EN}} & \multicolumn{2}{c}{\textbf{EN→TR}} \\
\textbf{Category} & OPET & NOPET & OPET & NOPET \\
\midrule
Death & 0.67 & 0.38  & 0.74 & 0.86 \\ 
Sexual Activity & 0.86 & 0.44 & 0.79 & 0.84 \\
Employment & 0.35 & 0.36 & 0.48 & 0.90 \\
Phys./Ment. Attr. & 0.79 & 0.59 & 0.68 & 0.70 \\
Politics & 0.23 & 0.18 & 0.50 & N/A \\
\bottomrule
\end{tabular}
\caption{Category-specific F1 scores for cross-lingual transfer. Models are trained on all source language OPETs (or all NOPETs), then tested on target language subsets, with F1 computed only for instances from each category.}
\label{tab:category}
\end{table}

Category-level analysis shows that transfer behavior is highly uneven across domains, with the largest degradations and gains, occurring in Employment and Politics. In the TR→EN direction (the challenging NOPET transfer), Politics (F1=0.18) and Employment (F1=0.36) show the weakest performance, as these categories contain English-specific PETs like between jobs and regime change that lack Turkish equivalents.

In contrast, EN→TR transfer remains robust even in the same categories (Employment F1=0.90, Death F1=0.86), demonstrating a substantial asymmetry at the category level. These results suggest that English-trained models exhibit more stable cross-category performance, even when target expressions lack direct equivalents, though this pattern may reflect resource imbalance and label distribution effects rather than deeper semantic generalization.

In the EN→TR direction, transfer to Turkish NOPETs often exceeds baseline while transfer to Turkish OPETs falls below baseline within the same category (e.g., Death: TR-NOPET at 0.86 exceeds its 0.68 baseline, while TR-OPET at 0.74 falls slightly above its 0.72 baseline; Employment: TR-NOPET at 0.90 far exceeds baseline, while TR-OPET at 0.48 falls well below baseline). This pattern reflects label distribution differences: Turkish NOPETs in these categories have higher proportions of euphemistic instances (e.g., Employment NOPETs: 82\% euphemistic vs OPETs: 44\% euphemistic), which aligns better with English-trained models' bias toward predicting euphemistic labels.

\subsection{Comparison with Zero-Shot Generative Models}
To contextualize our fine-tuning results, we evaluated GPT-4o \citep{openai2024gpt4o} using zero-shot prompting on all four splits. We prompted the model with a task description defining euphemisms and asked for binary classification (see Appendix~\ref{appendix:gpt4o} for the exact prompt).

\begin{table*}[t]
\centering
\begin{tabular}{lccccc}
\hline
\textbf{Model} & \textbf{EN OPET} & \textbf{EN NOPET} & \textbf{TR OPET} & \textbf{TR NOPET} & \textbf{Avg} \\
\hline
Frozen XLM-R (baseline) & 0.71 & 0.61 & 0.72 & 0.68 & 0.68 \\
\textbf{Fine-tuned XLM-R} & \textbf{0.85} & 0.62 & \textbf{0.80} & 0.79 & \textbf{0.77} \\
Zero-shot GPT-4o & 0.75 & \textbf{0.71} & 0.73 & \textbf{0.80} & 0.75 \\
\hline
\end{tabular}
\caption{In-domain performance comparison across approaches. Fine-tuned XLM-R results show models trained and tested on the same subset. Bold indicates best per column.}
\label{tab:model_comparison}
\end{table*}

Results reveal three key findings:

\textbf{1. Systematic prediction bias.} GPT-4o exhibits a consistent bias toward predicting "Euphemistic," with prediction rates ranging from 59.2\% (EN NOPETs) to 77.0\% (TR NOPETs), compared to actual label distributions of 49.4\%-59.0\% (Table~\ref{tab:gpt4o_bias}). This over-prediction is more pronounced for Turkish (+16-18 percentage points) than English (+10-12 percentage points), though the exact cause (e.g., pretraining data imbalance) cannot be verified without access to model internals.

\begin{table}[t]
\centering
\small
\begin{tabular}{lccc}
\toprule
\textbf{Split} & \textbf{Pred. \%} & \textbf{Label \%} & \textbf{$\Delta$} \\
\midrule
TR OPET  & 67.3 & 51.2 & +16.1 \\
TR NOPET & 77.0 & 59.0 & +18.0 \\
EN OPET  & 64.0 & 52.0 & +12.0 \\
EN NOPET & 59.2 & 49.4 & +9.8 \\
\bottomrule
\end{tabular}
\caption{GPT-4o prediction bias. $\Delta$ shows over-prediction of ``Euphemistic'' compared to actual label distribution.}
\label{tab:gpt4o_bias}
\end{table}

\textbf{2. Apparent NOPET advantage is an artifact.} GPT-4o achieves higher F1 on Turkish NOPETs (0.80) than OPETs (0.73), but this reflects alignment between its prediction bias (77\% euphemistic) and NOPET label distribution(59\% euphemistic) rather than meaningful linguistic generalization.

\textbf{3. Competitive but not superior to fine-tuning.} GPT-4o achieves an average F1 of 0.75 compared to fine-tuned XLM-R's 0.77, demonstrating that task-specific fine-tuning of a smaller encoder (278M parameters) can match or exceed zero-shot performance from much larger models.

Critically, GPT-4o's performance patterns differ fundamentally from XLM-R's transfer results. While XLM-R shows that OPET-trained models degrade less from baseline in the challenging TR→EN direction(reflecting semantic overlap constraining transfer), GPT-4o's apparent NOPET advantage is an artifact of prediction bias aligning with  label distributions rather than meaningful linguistic generalization. This distinction highlights that the OPET/NOPET framework is particularly informative for analyzing transfer dynamics in fine-tuned multilingual encoders, whereas zero-shot generative models exhibit different failure modes driven by prediction bias.


\section{Error Analysis}
\label{sec:qualitative}
To better understand the patterns driving these results, we analyzed representative errors in the zero-shot transfer predictions.

\subsection{The Cultural Gap (False Negatives)}
Transfer fails consistently when a euphemism relies on conceptual mappings absent in the source language. For example, in the TR $\to$ EN direction, the model failed to detect the English PET \textit{"between jobs"}:

\begin{quote}
\textit{Target Test (EN):} "I applied for temporary assistance when I was \textbf{between jobs} for a month to support my family."\\
\textit{Prediction:} Literal (False Negative).
\end{quote}

This PET for unemployment has no functional equivalent in Turkish, where direct terms are typically used. Similarly, the baseball-derived sexual activity PET \textit{"second base"} was consistently misclassified, as Turkish lacks this sports-to-intimacy mapping.

\subsection{Lexical Memorization (False Positives)}
Some errors suggest the model relies on specific tokens rather than context. This is evident in the Turkish term \textit{dört kollu} (literally: "four-armed"), a euphemism for a coffin. The model appears to have associated this term with euphemistic usage but failed to suppress the prediction in literal contexts.
\begin{quote}
\textit{Context:} "...\textbf{dört kollu} tanrıçanın tasviri..." \\
(\textit{...the depiction of the \textbf{four-armed} goddess...}) \\
\textbf{Prediction:} Euphemistic (False Positive).
\end{quote}
This pattern suggests that, in some cases, the model may be matching lexical items rather than interpreting context.

\subsection{Successful Cross-Lingual Transfer}
Transfer succeeds when both languages use similar expressions for the same taboo concept. For example, both English \textit{pass away} and Turkish \textit{vefat etmek} euphemize death through a departure framing:

\begin{quote}
\textbf{Source Training (EN):} "He \textit{passed away} last year."

\textbf{Target Test (TR):} "Geçen yıl \textit{vefat etti}." 
(Lit: "He passed away last year.")

\textbf{Prediction:} Euphemistic (True Positive).
\end{quote}
These cases involve OPETs where both languages express the same underlying concept with similar semantic framing, which may support transfer even without lexical overlap. 


\section{Conclusion and Future Work}
In this study, we investigated cross-lingual transfer in euphemism detection between English and Turkish by categorizing Potentially Euphemistic Terms (PETs) into semantically, pragmatically, and functionally overlapping (OPET) and non-overlapping (NOPET) subsets. Our experiments with XLM-R reveal three key findings:

First, overlap has limited impact when transferring from high-resource languages: models trained on English OPETs show minimal degradation from baseline on both Turkish OPETs (0.68 vs 0.72 baseline, a 0.04 drop) and Turkish NOPETs (0.70 vs 0.68 baseline, a slight gain), suggesting that extensive pretraining on English may provide more robust cross-lingual transfer regardless of overlap status in the target language. However, fine-tuning on English OPETs still underperforms the frozen baseline on Turkish OPETs, indicating that task-specific adaptation can reduce cross-lingual generalization.

Second, we observe a substantial transfer asymmetry most pronounced at the category level: English-to-Turkish transfer remains robust across domains (Employment F1=0.90, Death F1=0.86), while Turkish-to-English transfer degrades substantially (Employment F1=0.36, Death F1=0.38), with performance gaps exceeding 0.50 F1 points in some categories. This asymmetry is consistent with prior observations about resource imbalance in multilingual pretraining, though typological and morphological differences between English and Turkish cannot be disentangled in the present study.

Third, our error analysis reveals that successful transfer occurs when both languages use semantically similar expressions for the same taboo concept (e.g., death as departure in both English and Turkish), while failures stem from culture-specific mappings absent in the source language and, in some cases, apparent lexical memorization without contextual disambiguation.

For future work, we suggest: (1) extending the OPET/NOPET framework to additional language pairs with varying typological and resource characteristics (e.g., English-Spanish, Turkish-Azerbaijani), (2) investigating whether the observed asymmetry persists across different model architectures (e.g., mBERT, mT5), and (3) developing training strategies that reduce lexical memorization and improve context-dependent classification.


\section*{Limitations}

\paragraph{Language Pair Selection.}
Our analysis was restricted to English and Turkish, which differ typologically (analytic vs agglutinative) and in pretraining resource availability. While this contrast enables us to study resource asymmetry, it limits generalizability. The observed transfer patterns may reflect typological distance, morphological complexity, or other language-specific factors beyond resource imbalance. Future research should extend the OPETs/NOPETs framework to multiple language pairs with varying typological and resource characteristics (e.g., English-Spanish, Turkish-Azerbaijani, Chinese-Japanese) to disentangle these factors.

\paragraph{Sociolinguistic Variation}
Euphemistic language is shaped not only by typological differences between languages but also by social and cultural factors. While English and Turkish differ typologically, euphemisms additionally vary across regions, dialects, and speaker communities, including varieties of Turkish spoken outside Turkey. In this study, we focus exclusively on standard Turkey Turkish as represented in web-based corpora, and do not explicitly model regional or diasporic variation. As a result of these factors, both detection performance and cross-lingual transfer may be affected.

\paragraph{Category-Level Data Sparsity.}
Semantic categories were inherently imbalanced, with some severely underrepresented in specific subsets. For example, Politics contains only 1 Turkish OPETs with 11 instances, compared to 21 Death OPETs with 661 instances. While we report category-specific performance (Table~\ref{tab:category}), findings for low-frequency categories (Politics, Illness, Substances) should be interpreted cautiously due to limited statistical power. The absence of Turkish Politics NOPETs (N/A in Table~\ref{tab:category}) prevents evaluation in this category. Similarly, very small categories like Turkish Politics OPETs (11 instances) exhibit high variance.

\paragraph{Category-Level Sample Size.}
Some semantic categories contain very few instances in specific OPETs/NOPETs subsets (e.g., Turkish Politics OPETs: 11 instances, English Death NOPETs: 52 instances). With 10-fold cross-validation, individual folds may contain as few as 1-5 test instances, leading to high variance in category-specific F1 scores (Table~\ref{tab:category}). These results should be interpreted cautiously.

\paragraph{Binary Classification Scope.}
Our study focused on binary classification (Euphemistic vs Literal), which does not capture gradations of euphemistic strength or speaker intent. Additionally, our experimental setup does not explicitly measure the model's ability to leverage broader context for disambiguation. Future work could investigate whether larger context windows, attention analysis, or multi-task learning (e.g., predicting both label and confidence) improve performance on pragmatically ambiguous cases.

\paragraph{Single Model Architecture.}
We evaluated only XLM-R (with brief GPT-4o comparison), limiting our ability to determine whether findings generalize across architectures. Future work should compare multiple multilingual encoders (e.g., mBERT, mT5, BLOOM) to assess whether the observed patterns reflect XLM-R-specific biases or general cross-lingual transfer dynamics.

\paragraph{Training Instability in NOPETs.}
Additionally, models trained on English NOPETs exhibited high variance, reflecting the diverse nature of non-overlapping euphemisms, which lack consistent cross-lingual patterns and employ varied linguistic strategies. This instability contrasts with the lower variance observed in OPET-trained models.

 \paragraph{Prompt Language Sensitivity.} Our comparison with GPT-4o relies on zero-shot prompting with language-matched instructions: English prompts were used for English data and Turkish prompts for Turkish data (Appendix~\ref{appendix:gpt4o}). We did not conduct a controlled ablation comparing Turkish sentences prompted in English versus Turkish. While language-matched prompting is a reasonable choice for pragmatic phenomena such as euphemisms, differences in prompt language may influence model sensitivity to culturally embedded or idiomatic expressions. Future work could explicitly evaluate the effect of prompt language on classification behavior for low-resource or culturally specific euphemisms.

\section*{Acknowledgments}
This research was supported in part by the National Science Foundation under Grant No.~\textit{2226006}. We thank Ecem Küçük and Mihriban Kandemir for their contributions to the manual annotation of the Turkish PETs dataset.

\bibliography{custom}
\bibliographystyle{acl_natbib}

\raggedbottom

\pagebreak

\appendix

\section{Category Statistics}
\label{sec:appendix}

\begin{table*}[t]
\centering
\small
\setlength{\tabcolsep}{8pt} 
\renewcommand{\arraystretch}{1.2}
\caption{Category statistics for Turkish (TR) and English (EN) PETs OPETs and NOPETs subsets.}
\label{tab:category_statistics}
\begin{tabular}{l|cc|cc|cc|cc}
\toprule
\textbf{Category} & \multicolumn{2}{c|}{\textbf{TR OPETs}} & \multicolumn{2}{c|}{\textbf{TR NOPETs}} & \multicolumn{2}{c|}{\textbf{EN OPETs}} & \multicolumn{2}{c}{\textbf{EN NOPETs}} \\
 & \textbf{PETs} & \textbf{Instances} & \textbf{PETs} & \textbf{Instances} & \textbf{PETs} & \textbf{Instances} & \textbf{PETs} & \textbf{Instances} \\
\midrule
Death & 16 & 484 & 5 & 177 & 14 & 528 & 1 & 52 \\
Body Functions/Parts & 5 & 290 & 9 & 179 & 3 & 27 & 1 & 5 \\
Sexual Activity & 4 & 134 & 4 & 92 & 5 & 118 & 5 & 147 \\
Employment/Finances & 5 & 122 & 2 & 45 & 8 & 292 & 6 & 138 \\
Physical/Mental Attributes & 5 & 89 & 6 & 168 & 5 & 132 & 14 & 421 \\
Politics & 1 & 11 & 0 & 0 & 1 & 19 & 2 & 64 \\
Illness & 0 & 0 & 2 & 65 & 0 & 0 & 1 & 20 \\
Substances & 0 & 0 & 1 & 80 & 0 & 0 & 4 & 120 \\
Miscellaneous & 0 & 0 & 5 & 198 & 0 & 0 & 1 & 32 \\
\bottomrule
\end{tabular}
\end{table*}

\section{Annotation Guidelines for OPETs and NOPETs}
\label{sec:appendix_guidelines}
This appendix describes the guidelines provided to annotators for classifying Potentially Euphemistic Terms (PETs) as either Overlapping PETs (OPETs) or Non-Overlapping PETs (NOPETs). The annotation scheme is based on \textbf{functional, semantic, and pragmatic equivalence} between euphemistic expressions across languages.

\subsection{Definitions}

\paragraph{Euphemisms.}
Euphemisms are mild or indirect expressions used in place of harsher, offensive, or taboo terms. They are commonly used for politeness when discussing sensitive topics (e.g., \textit{passed away} instead of \textit{died}) or to obscure unpleasant or socially sensitive realities (e.g., \textit{enhanced interrogation} instead of \textit{torture}).

\paragraph{Potentially Euphemistic Terms (PETs).}
PETs are words or phrases that can be used euphemistically in some contexts but may also have literal, non-euphemistic meanings in others. For example, the term \textit{dismissed} is a PET because it can euphemistically mean ``fired'' or literally mean ``rejected'' or ``ignored,'' depending on context.

\subsection{Classification Criteria}

\paragraph{Non-Overlapping PETs (NOPETs).}
A PET is classified as a NOPET if it exists in the source language but has no euphemistic equivalent in the target language that refers to the same underlying taboo concept.
\begin{itemize}
    \item \textbf{Example:} The English PET \textit{the birds and the bees}, which euphemistically refers to sex education, has no lexical or functional euphemistic equivalent in Turkish. It is therefore classified as a NOPET.
\end{itemize}

\paragraph{Overlapping PETs (OPETs).}
A PET is classified as an OPET if both languages contain expressions that can function euphemistically for the same underlying taboo concept, regardless of whether the expressions are literal translations of one another.
\begin{itemize}
    \item \textbf{Criteria:} Both languages must have expressions that \emph{can} euphemize the same taboo concept.
    \item \textbf{Example:} \textit{Pass away} (English) and \textit{vefat etmek} (Turkish) are classified as OPETs because both are used to euphemistically replace the concept of \textit{death}.
\end{itemize}

\subsection{Decision Procedure for Annotators}

Annotators were instructed to follow the decision procedure below:
\begin{enumerate}
    \item \textbf{Identify the taboo concept:} Determine the specific taboo or sensitive concept the PET replaces in Language~A (e.g., death, sexual activity, illness).
    \item \textbf{Search the target language:} Identify whether Language~B contains any expression that euphemistically refers to the same taboo concept.
    \item \textbf{Evaluate equivalence:}
    \begin{itemize}
        \item If a euphemistic expression exists in Language~B that refers to the same taboo concept (even without lexical overlap), classify the PET as \textbf{OPET}.
        \item If no such euphemistic expression is found after a thorough search, classify the PET as \textbf{NOPET}.
    \end{itemize}
\end{enumerate}

\subsection{Handling Usage Asymmetries}

Differences in usage frequency or obligatoriness do not affect OPET classification. The annotation focuses on \textbf{euphemistic potential} rather than usage distribution.
\begin{quote}
    \textbf{Rule:} If a term in Language~A can be used both euphemistically and literally (e.g., \textit{pass away}), while its counterpart in Language~B is used only euphemistically (e.g., \textit{vefat etmek}), the pair is still classified as OPET, provided that both expressions can euphemize the same taboo concept.
\end{quote}

\subsection{Examples}

Table~\ref{tab:appendix_examples} presents representative examples of OPET and NOPET classifications.

\begin{table*}[t!] 
\centering
\small
\caption{Examples of OPET and NOPET classifications. While lexical forms may differ, OPETs share a functional role in euphemizing the same taboo concept.} 
\label{tab:appendix_examples}
\renewcommand{\arraystretch}{1.2}
\begin{tabular}{l l l c p{5cm}}
    \toprule
    \textbf{English PET} & \textbf{Turkish PET} & \textbf{Taboo Concept} & \textbf{Label} & \textbf{Notes} \\
    \midrule
    comfort woman & hayat kadını & Sex work & OPET & Different expressions, same euphemistic function. \\
    pass away & vefat etmek & Death & OPET & Functional equivalence across languages. \\
    the birds and the bees & --- & Sex education & NOPET & Non-Overlap. \\
    --- & fındık kırmak & Sexual activity & NOPET & Turkish slang; no English equivalent. \\
    \bottomrule
\end{tabular}
\end{table*}

\section{GPT-4o Experimental Details}
\label{appendix:gpt4o}

To contextualize our fine-tuning results, we evaluated GPT-4o \citep{openai2024gpt4o} using zero-shot prompting on all four splits. We prompted the model with a task description defining euphemisms and asked for binary classification. The exact prompts used are shown below.

\subsection{Prompt Templates}

For each instance, we provided a system prompt defining the task and a user prompt containing the sentence and target term.

\paragraph{System Prompt (English):}
\begin{quote}
\texttt{You are a linguistics expert analyzing English text for euphemistic language. A euphemism is an indirect expression that softens harsh, offensive, or taboo concepts like death, illness, firing, bodily functions, or sexual activity. Respond with ONLY one word: "Euphemistic" or "Literal".}
\end{quote}

\paragraph{System Prompt (Turkish):}
\begin{quote}
\texttt{You are a linguistics expert analyzing Turkish text for euphemistic language. A euphemism is an indirect expression that softens harsh, offensive, or taboo concepts like death, illness, firing, bodily functions, or sexual activity. Respond with ONLY one word: "Euphemistic" or "Literal".}
\end{quote}

\paragraph{User Prompt Template:}
\begin{quote}
\texttt{Analyze this [English/Turkish] sentence:}\\[0.5em]
\texttt{Sentence: "\{sentence\}"}\\
\texttt{Term: "\{term\}"}\\[0.5em]
\texttt{Question: Is the term "\{term\}" used as a EUPHEMISM (indirect, softening expression) or LITERALLY (direct meaning)?}\\[0.5em]
\texttt{Classification:}
\end{quote}

\subsection{Inference Parameters}

We used the following parameters for GPT-4o inference:
\begin{itemize}
    \item \textbf{Model:} \texttt{gpt-4o}
    \item \textbf{Temperature:} 0 (deterministic output)
    \item \textbf{Max tokens:} 20
\end{itemize}

\subsection{Response Parsing}

Responses were parsed case-insensitively to extract classifications. A response containing ``euphemistic'' or ``euphemism'' was labeled as Euphemistic; responses containing ``literal'' were labeled as Literal. Responses containing refusal phrases (e.g., ``cannot,'' ``I apologize'') or unexpected outputs were logged separately. Across all splits, refusals and unexpected responses constituted less than 1\% of total predictions.

\end{document}